\documentclass[conference]{IEEEtran}
\IEEEoverridecommandlockouts
\usepackage{cite}
\usepackage{amsmath,amssymb,amsfonts}
\usepackage{algorithmic}
\usepackage{graphicx}
\usepackage{hyperref}
\usepackage{textcomp}
\usepackage{xcolor}
\usepackage{listings}

\newcommand{\change}[1]{{\color{black}#1}}

\definecolor{codegreen}{rgb}{0,0.6,0}
\definecolor{codegray}{rgb}{0.5,0.5,0.5}
\definecolor{codepurple}{rgb}{0.58,0,0.82}
\definecolor{backcolour}{rgb}{0.95,0.95,0.92}

\lstdefinestyle{mystyle}{
    backgroundcolor=\color{backcolour},   
    commentstyle=\color{codegreen},
    keywordstyle=\color{magenta},
    numberstyle=\tiny\color{codegray},
    stringstyle=\color{codepurple},
    basicstyle=\ttfamily\footnotesize,
    breakatwhitespace=false,         
    breaklines=true,                 
    captionpos=b,                    
    keepspaces=true,                 
    numbers=left,                    
    numbersep=5pt,                  
    showspaces=false,                
    showstringspaces=false,
    showtabs=false,                  
    tabsize=2
}

\def\BibTeX{{\rm B\kern-.05em{\sc i\kern-.025em b}\kern-.08em
    T\kern-.1667em\lower.7ex\hbox{E}\kern-.125emX}}

\def\modelname{\textsc{HLAT}}
\def\trn1{\emph{trn1}}
\def\trainium{\textsc{Trainium}}

\usepackage{subcaption}
\usepackage{multirow}
\usepackage{booktabs}

\usepackage{lipsum}

\begin{document}
\bstctlcite{IEEEexample:BSTcontrol}

\title{HLAT: High-quality Large Language Model Pre-trained on AWS Trainium}

\author{Haozheng Fan$^{*1}$\thanks{$^{*}$Both authors contributed equally to this research.}
, Hao Zhou$^{*2}$, Guangtai Huang$^1$, Parameswaran Raman$^1$, Xinwei Fu$^1$, \\
Gaurav Gupta$^2$, Dhananjay Ram$^3$, Yida Wang$^1$, Jun Huan$^2$ \\
{$^1$Amazon Web Services, $^2$AWS AI Labs, $^3$AGI Foundations, Amazon}
\\
  {{\{fanhaozh,\hspace{0.2em}zhuha,\hspace{0.2em}guangtai,\hspace{0.2em}prraman,\hspace{0.2em}fuxinwe,\hspace{0.2em}gauravaz,\hspace{0.2em}radhna,\hspace{0.2em}wangyida,\hspace{0.2em}lukehuan\}}@amazon.com}
}

\maketitle

\begin{abstract}
Getting large language models (LLMs) to perform well on the downstream tasks requires pre-training over trillions of tokens. This typically demands a large number of powerful computational devices in addition to a stable distributed training framework to accelerate the training. 
The growing number of applications leveraging AI/ML led to a scarcity of the expensive conventional accelerators (such as GPUs), which emphasizes the need for the alternative specialized-accelerators that are \textit{scalable and cost-efficient}. 
AWS \trainium~is the second-generation machine learning accelerator purposely built for training large deep learning models.
However, training LLMs with billions of parameters on AWS \trainium~is challenging due to its relatively nascent software ecosystem. 
In this paper, we showcase \modelname: a family of 7B and 70B decoder-only LLMs pre-trained using 4096 AWS \trainium~accelerators over 1.8 trillion tokens. The performance of \modelname~is benchmarked against popular open source models including LLaMA and OpenLLaMA, which have been trained on NVIDIA GPUs and Google TPUs, respectively. On various evaluation tasks, we show that \modelname~achieves model quality on par with the baselines of similar model size. We also open-source all the training scripts and configurations of \modelname\footnote{{https://github.com/awslabs/HLAT}} and share the best practice of using the NeuronX Distributed Training (NxDT), a customized distributed training library for AWS \trainium. Our work demonstrates that AWS \trainium~powered by NxDT is able to successfully pre-train state-of-the-art LLM models with high performance and cost-effectiveness.
\end{abstract}



\section{Introduction}
Large language models (LLMs), based on transformer architecture \cite{vaswani2017attention} and trained on massive text data, are the most recent breakthrough in artificial intelligence. They not only show remarkable capabilities in understanding and generating text \cite{li2022pretrained}, but offer immense potential across diverse downstream tasks, such as machine translation \cite{hendy2023good}, information retrieval \cite{zhu2023large}, code generation \cite{roziere2023code} and so on \cite{LLMSurvey}. 

Pre-training is the crucial first step in building LLMs because it lays the foundation for their impressive capabilities. It initializes the model with random weights, and trains the model to convergence using tokens from a large text corpus. The training process is designed to be self-supervised. For decoder-only models, such as GPT \cite{gpt3} and LLaMA \cite{llama1,llama2,llama3}, the model is trained to predict the next token in a given sequence. Eventually, the model learns everything ranging from syntax and semantics to world knowledge and commonsense reasoning with a large amount of training data. 
Pre-training provides the raw material - the language skills and understanding, which facilitates subsequent fine-tuning for various downstream tasks.

Since pre-training requires a large amount of training data (trillions of tokens), it demands highly on computational resources. 
Advanced AI accelerators, such as AWS \trainium\footnote{{https://aws.amazon.com/machine-learning/trainium}}, Google TPU, and NVIDIA A100/H100 GPUs, have been specifically designed for such workloads. 
These AI accelerators are often integrated with dedicated tensor processing units  which offer fast matrix operations and high training throughput. 
They also have much larger on-chip memory (tens of GBs per accelerator)
and high communication bandwidth (hundreds of Gbps) between accelerators across different machines,
which allows pre-training of larger models with efficient hardware utilization. 

Even with the powerful AI accelerators, due to the sheer size and complexity of LLMs, it's impractical to train them on a single device which has limited memory and processing power to handle the massive datasets, model parameters, and intricate calculations involved in LLM training. 
Practitioners rely on distributed training libraries~\cite{rasley2020deepspeed,shoeybi2020megatronlm,FairScale2021,zhao2023pytorch} to orchestrate a number of accelerators to conduct the training together.
Distributed libraries can shard the model parameters and optimizer states across multiple accelerators with different kinds of parallelism strategy, allowing training of models with multi-billion parameters.
They also spread the workload across multiple machines, effectively tapping into a combined pool of resources, to significantly reduce the training time. 

Although there have been many successful demonstrations of pre-training LLMs on conventional accelerators (GPUs and TPUs) using state-of-the-art distributed training libraries \cite{rasley2020deepspeed,shoeybi2020megatronlm,FairScale2021,zhao2023pytorch},
training LLMs with billions of parameters on AWS \trainium~is still challenging. 
First, \trainium~uses a relatively nascent software ecosystem ranging from runtime, compiler, to distributed training library. 
The training script developed for other accelerators needs to be adjusted to comply with the low-level APIs and operators supported by \trainium. 
Second, the optimal training configurations that ensure stable convergence and optimal training throughput may also differ from other accelerators, such as level of precision, dimensions of 3D parallelism, compiler flags and so on.
On the other hand, Amazon EC2 \trn1 instance, equipped with AWS \trainium~accelerators, provides the comparable computation power to Amazon EC2 \emph{p4d} instance, equipped with Nvidia A100 40GB GPUs, but comes with only $\sim$60\% of the price. This makes it appealing to fully utilize the compute power of AWS \trainium~for LLM pre-training.

\change{In this paper, we make the following contributions:
\begin{itemize}\setlength{\itemsep}{-0pt}
    \item Our work is the first to reproduce a SOTA LLM model on a completely new hardware - AWS \trainium. We end-to-end pre-train \modelname-7B and \modelname-70B (\textbf{H}igh-quality \textbf{L}LM pre-trained on \textbf{A}WS \textbf{T}\textsc{rainium}), following the architecture described in \cite{llama2} from scratch. The pre-training covers 1.8 trillion tokens and is performed on up to 256 Amazon EC2 \emph{trn1.32xlarge} instances with totalling up to 4096 AWS \trainium~accelerators. 
    \item We evaluate and show that both \modelname-7B and \modelname-70B perform comparable to models of similar size trained on other AI accelerators including LLaMA and OpenLLaMA. The evaluation is performed over a variety of tasks including commonsense reasoning, world knowledge, MMLU\cite{hendrycks2021ethics}, math, coding, etc. 
    \item We propose multiple techniques to improve training efficiency on AWS \trainium~such as a novel online dataloader, layer coalesing, selective activation checkpointing, precision strategy, and fault recovery mechanisms. These techniques save significant time and compute resources for pretraining on large datasets, and can be applied to other accelerators (GPU, TPU, etc.) as well. 
    \item We open-source all the training scripts and configurations of \modelname-7B and \modelname-70B models\footnote{{https://github.com/awslabs/HLAT}}, including the Pytorch training script, model definition script, and training configuration files (including all hyper-parameters). We also share best practices of pre-training on AWS \trainium~and NeuronX Distributed Training (NxDT), such as sharding strategies, training precisions, compiler settings, etc. With those artifacts, practitioners can easily reproduce \modelname~and pre-train their own custom LLMs on AWS \trainium. 
\end{itemize}
}

\section{Background - Distributed Training on AWS \trainium}
\label{sec:back}
\textbf{AWS \trainium} is the second-generation machine learning accelerator that AWS purposely built for deep learning training. Each \trainium~accelerator includes two NeuronCores. Each NeuronCore has 16 GB of high-bandwidth memory, and delivers up to 95 TFLOPS of FP16/BF16 compute power. 
In this study, we trained our model on Amazon EC2 \emph{trn1.32xlarge} instances: each instance is equipped with 16 \trainium~accelerators, and supports 800 Gbps intra-instance network bandwidth through NeuronLink.
The aggregating compute power of Amazon EC2 \emph{trn1.32xlarge} is 3040 TFLOPS in FP16/BF16, slightly higher to its GPU instance counterpart Amazon EC2 \emph{p4d.24xlarge} at 2496 TFLOPS, but at a much lower price (\emph{trn1.32xlarge} \$21.50 vs. \emph{p4d.24xlarge} \$32.77).

\textbf{AWS Neuron} is a software development kit (SDK)\footnote{{https://github.com/aws-neuron/aws-neuron-sdk}} with a compiler, runtime, and profiling tools that unlocks high-performance and cost-effective deep learning acceleration on AWS \trainium. Neuron is natively integrated with PyTorch \cite{zhao2023pytorch} and TensorFlow \cite{tensorflow2015-whitepaper}, and offers features such as FP32 autocasting, stochastic rounding, collective communication, custom operators, and so on. 

\textbf{NeuronX Distributed Training} (NxDT)\footnote{{https://awsdocs-neuron.readthedocs-hosted.com/en/latest/libraries/nxd-training/overview.html}}, as part of Neuron SDK, is developed to enable high-efficiency distributed training on \trainium: 
NxDT supports a variety of distributed training techniques, such as 3D parallelism~\cite{shoeybi2020megatronlm}, i.e., Tensor Parallelism (TP), Pipeline Parallelism (PP) and Data Parallelism (DP). 
To reduce the activation memory during training, activation checkpointing~\cite{chen2016training} and sequence parallelism \cite{korthikanti2022reducing} are naturally supported with the 3D parallelism.
NxDT also supports Zero Redundancy Optimizer Stage 1 (ZeRO-1)~\cite{rajbhandari2020zero} to shard optimizer states, which can be applied simultaneously with 3D parallelism.
NxDT provides unified interfaces to port custom models and run training scripts on AWS \trainium. To train models from Huggingface \texttt{transformers} library on \trainium~accelerators with NxDT, it only requires simple code changes in certain layers of the model.
NxDT supports TP with mixed degrees, i.e., users can use more than one TP degrees to shard different model parameters.
This is helpful when some model parts of LLMs are not compatible with a unified large TP degree, e.g., with Grouped Query Attention (GQA)~\cite{hudson2019gqa}.
Finally, NxDT supports automatic fault recovery and checkpointing. In case of hardware failures or communication timeouts, NxDT can automatically restart training from latest auto-saved checkpoints without manual intervention, which is critical for maintaining system uptime and training efficiency. 

\section{Method}
\subsection{Model Architecture and Hyperparameters}
\label{ssec:train_hyper}
\modelname~models adopt the decoder-only transformer architecture and apply same modifications used in LLaMA \cite{llama1, llama2, llama3}, including pre-normalization with RMSNorm, SwiGLU activation function, and Rotary Embeddings. \modelname-70B in addition applies GQA \cite{hudson2019gqa} with group size of 8. The models are trained with a maximum sequence length of 4096. 

We adopt training hyperparameters used in LLaMA2 \cite{llama2}. Specifically, the global batch size is 1024 sequences, so each step covers about 4 million tokens. We use a cosine learning rate scheduler. The maximum learning rate is $3e^{-4}$ for \modelname-7B, and $1.5e^{-4}$ for \modelname-70B. The minimum learning rate decays to $10\%$ of maximum learning rate. We use a linear warmup of 2000 steps. The overall learning rate scheduler is plotted in Figure \ref{fig:lr}. We use AdamW optimizer with $\beta_1=0.9$ and $\beta_2=0.95$. We use weight decay value of 0.1 for all parameters, including normalization weights. Gradient-norm clipping of 1.0 is applied for training stability. 

\subsection{Training Dataset and Dataloader}\label{sec:dataloader}
\change{Our pre-training dataset includes RedPajama-1T \cite{together2023redpajama}, peS2o \cite{peS2o}, and OpenWebMath \cite{paster2023openwebmath}. \modelname-7B is purely trained on RedPajama-1T with 1.8 trillion tokens. \modelname-70B is initially trained on RedPajama-1T over 1.4 trillion tokens and is then continually trained on an up-sampled dataset with RedPajama-1T, peS2o and OpenWebMath for 400B tokens (see Section \ref{sec:upsampling} for details). }

We designed a novel dataloader which performs both tokenization and packing \textit{online} during training. The dataloader takes one or more dataset files in Apache Arrow format \cite{arrow2020}. All samples are randomly shuffled and split into several subsets according to the total DP ranks. Each data split is treated as an independent data stream. For training efficiency, we use sample concatenation, i.e., if a sample is shorter than the maximum sequence length of the model, we concatenate it with the following sample(s) to curate a sequence with total length equal or more than maximum sequence length. Any left over tokens from current concatenated sequence is used in the following sequence. The samples within a concatenated sequence is concatenated with a special end of sentence (EOS) token. This gives the model necessary information to infer that the text separated by EOS token are unrelated~\cite{gpt3}. Note that samples in the same concatenated sequence may be from very different sources or can be of different formats (e.g. natural language and codes). Finally, each batch of sequences are tokenized on the fly during the training. 

The online tokenization has no impact on training throughput as the tokenization for future samples/batch happens during forward-backward pass of current samples/batch - we use CPU for tokenization and \trainium~devices for training, so the computations are in parallel. In comparison, the offline dataloader requires pre-tokenization of entire datasets which costs a lot of developer time and compute resources for large datasets. The offline Nemo dataloader\cite{Harper_NeMo_a_toolkit}, for example, takes about 154 hours to pre-tokenize a dataset with 2 trillion tokens.

\subsection{Orchestration}
\modelname~pre-training is performed on clusters with 256 \emph{trn1.32xlarge} instances (nodes), totalling to 4096 AWS \trainium~accelerators. Both Amazon EKS and AWS ParallelCluster can effectively manage the training cluster. Accelerators within same node are connected with NeuronLink\footnote{{https://awsdocs-neuron.readthedocs-hosted.com/en/latest/general/arch/neuron-hardware/trainium.html}}. The nodes within the cluster are interconnected through Elastic Fabric Adapter (EFA)\footnote{https://aws.amazon.com/hpc/efa/}. EFA is a network interface with uniquely designed operating system that bypasses traditional hardware interfaces, significantly enhancing performance for inter-node communications, a critical factor for collectives operations in distributed training. 

\subsection{Training Efficiency}
LLaMA \cite{llama1} model uses the efficient implementation features for pre-training on GPUs, that include \texttt{xformer} library, activation checkpointing, model parallelism, and computation/communication overlapping, etc. Similar features are also supported by \trainium~and Neuron SDK, as well as some unique enhancement such as BF16 with stochastic rounding. Below, we list the key features and configurations used in our model pretraining to improve the efficiency. 

\textbf{Model Parallelism}: HLAT-7B is pre-trained over 64 nodes with TP=8, PP=1, and DP=256. \modelname-70B is pre-trained over 256 nodes with TP=32, PP=8, and DP=32. 
Both use sequence parallel (SP). This sharding configuration is observed to have the highest throughput.

\textbf{Selective Activation Checkpointing}: We use selective activation checkpointing \cite{korthikanti2022reducing} to improve the training efficiency. It has slightly higher memory cost as full activation checkpointing, but increases the overall training throughput. 

\change{\textbf{Training precision}: Pre-training with full precision (FP32) is in-efficient for large LLMs, but generic half-precision training (BF16 or FP16) often has numerical stability issues \cite{micikevicius2017mixed}. On GPU, mixed precision training~\cite{micikevicius2017mixed} is widely used to achieve similar precision as full FP32 with better efficiency. For \modelname-7B, we used BF16 with stochastic rounding (SR)~\cite{sr}, featured by AWS \trainium. Stochastic rounding, which theoretically provides an unbiased estimate of the input, prevents the computation precision-loss in BF16 by performing the rounding operations in a probabilistic manner. Empirically, we found that BF16 with SR shows the same convergence behavior as mixed precision training for \modelname-7B, with higher training throughput and lower memory footprint. However, on \modelname-70B, BF16 with SR shows worse convergence than mixed precision training. 
{We found that mainly due to the nondeterministic normalization error introduced by stochastic rounding (see Section \ref{sec:sr_err}), and low-precision computations errors in gradient and attention operations. Such errors accumulate and signify on large models with more parameters and deeper layers.}
Therefore, we developed a standard mixed precision training strategy. Specifically, unless specified, all computation and storage use BF16 without stochastic rounding. This strategy uses FP32 in precision sensitive operators, local gradient accumulation, and global gradient synchronization. It also uses master weights and FP32 optimizer states. Finally, it uses ZeRO-1 \cite{rasley2020deepspeed} for memory efficiency.}

\textbf{Constant Attention Mask}: 
As a decoder-only model, \modelname~pre-training uses a constant attention mask (lower-triangular) matrix. Instead of passing attention mask as an input tensor in model training, AWS \trainium~supports creating attention masks on accelerators directly before use. The masked-out tensors will not be computed in the first place, which avoids redundant computation, saves host memory usage, and increases training throughput. To enable this feature in training script, the attention mask tensor is directly defined using \texttt{torch.triu} function and mapped to \texttt{device='xla'}. The Neuron compiler will therefore enable the optimization during compilation. 

\textbf{Coalescing Layers with Same Inputs}: 
We coalesced linear layers with the same inputs to reduce the communication in tensor and sequence parallelism, and increase efficiency of matrix operations. Specifically, the \texttt{Q,K,V} layers in an attention block are coalesced, and the two linear projections layers in SwiGLU \cite{shazeer2020glu} are also coalesced. Listing \ref{qkv} shows a code snippet for implementation. Note that this technique also applies to other accelerators such as GPU and TPU. 

\begin{lstlisting}[language=Python, caption=Example of layer coalescing., label={qkv}] 
# Without layer coalescing
query_states = q_proj(hidden_states) # hidden_size
key_states = k_proj(hidden_states)   # hidden_size
value_states = v_proj(hidden_states) # hidden_size

# With layer coalescing
qkv_states = qkv_proj(hidden_states) # 3*hidden_size
query_states, key_states, value_states = qkv_states.split(3, dim=2)   
\end{lstlisting}

\textbf{Compiler Optimization}: we use compiling flag \\ \texttt{-{}-distribution-strategy=llm-training} and \texttt{-{}-model-type=transformer} to enable the compiler to perform optimizations applicable to LLM (transformer model) training runs that shard parameters, gradients, and optimizer states across data-parallel workers. We set Neuron environment variable \texttt{NEURON\_FUSE\_SOFTMAX=1} to enable compiler optimizations on custom lowering for Softmax operation. Finally, we used {\small{\texttt{NEURON\_RT\_ASYNC\_EXEC\_MAX\_INFLIGHT\_REQUESTS=3}}} to reduce training latency with asynchronous execution. This overlaps some executions of accelerators and host (CPU). 

\section{Training Process}
\subsection{Training Curves}
\begin{figure*}[!ht]
    \centering
    \begin{minipage}{0.5\columnwidth}
    \centering
        \includegraphics[width=0.95\linewidth]{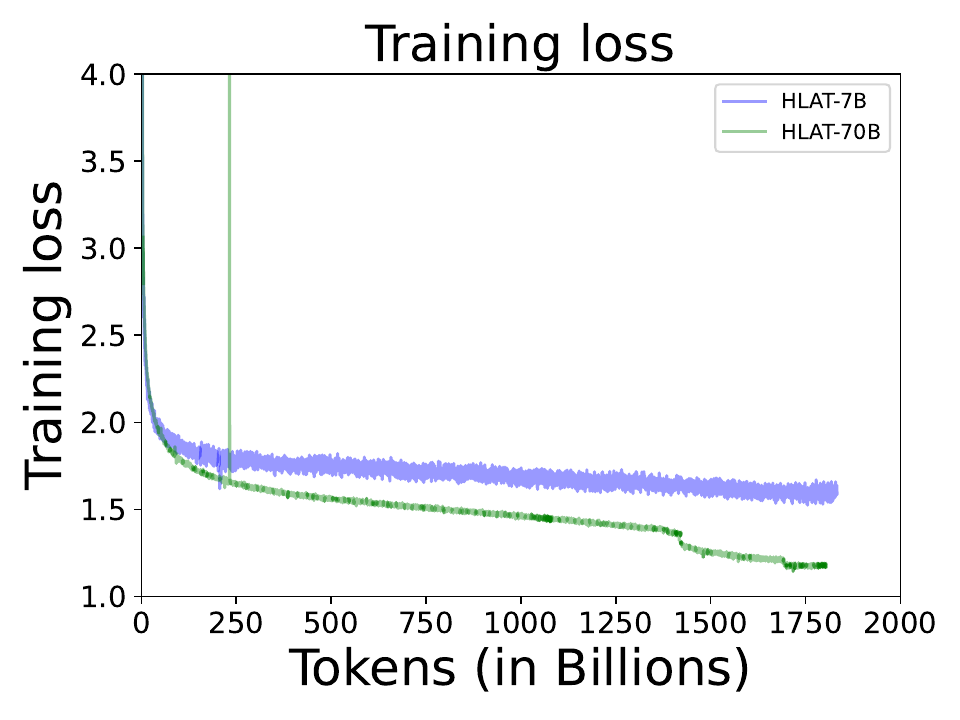}
        \subcaption{}
        \label{fig:loss}
    \end{minipage}
    \begin{minipage}{0.5\columnwidth}
    \centering
        \includegraphics[width=0.95\linewidth]{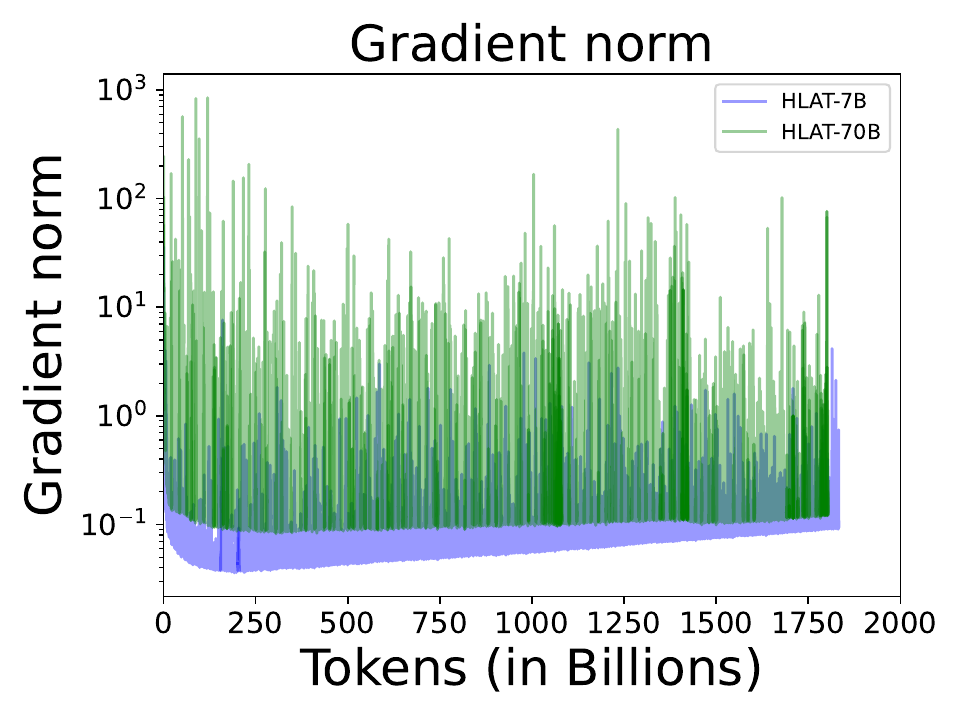}
        \subcaption{}
        \label{fig:gradient_norm}
    \end{minipage}
    \begin{minipage}{0.5\columnwidth}
    \centering
        \includegraphics[width=0.95\linewidth]{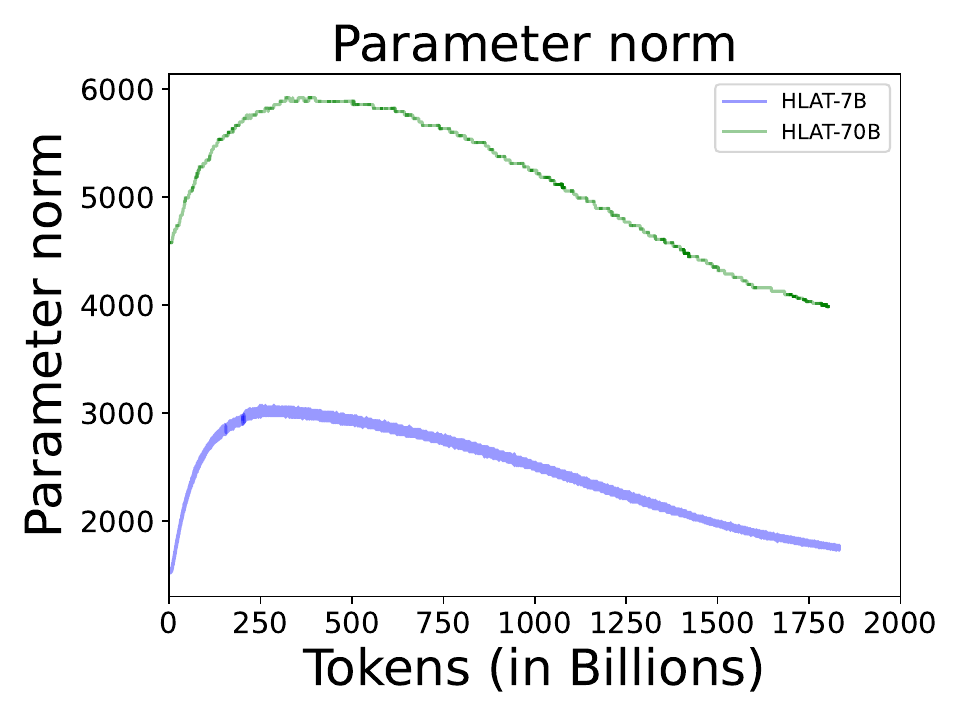}
        \subcaption{}
        \label{fig:parameter_norm}
    \end{minipage}
    \begin{minipage}{0.5\columnwidth}
    \centering
    \includegraphics[width=0.95\linewidth]{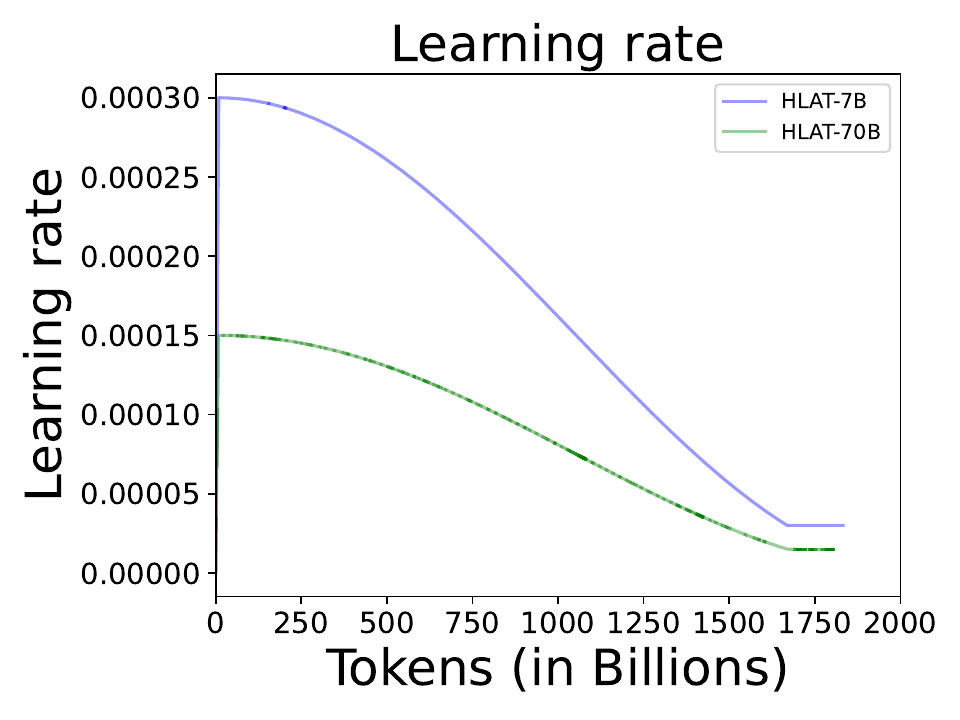}
    \subcaption{}
    \label{fig:lr}
    \end{minipage}
    \caption{\modelname~training progress. (a) The training loss vs number of tokens (in billions) seen by the model during training. Gradient/Parameter norm vs number of tokens in (b)/(c), respectively. (d) The learning rate schedule vs number of tokens. The warm-up steps are 2000 iterations (about 8 billion tokens, see Section\,\ref{ssec:train_hyper}).}
    \label{fig:training_curves}
\end{figure*}

During the training process, we monitor the cross entropy training loss,
as well as $l_2$ norm of gradients and $l_2$ norm of parameters for debugging training stability. Figure \ref{fig:loss} shows the training loss over global batches, reduced over all data parallel ranks. The training loss decreases rapidly over the initial $\sim$250 billion tokens, and enters a log-linear decrease afterwards. Similar trends are observed in other LLM pre-training \cite{llama1,llama2,openlm2023openllama}. 

In Figure \ref{fig:gradient_norm}, we show the gradient $l_2$ norm during the training. Overall, we see that the gradient norm is stable across the training journey without divergence. Note that gradient spikes are common in LLM pre-training when using layer-normalization, or even RMSNorm \cite{takase2023spike}, and sometimes due to overflow in low-precision, such as 16-bit floats. We show an assuring trend in Figure\,\ref{fig:gradient_norm} even with using 16-bit floats. 

\begin{figure}[h]
  \centering
  \includegraphics[width=.85\linewidth]{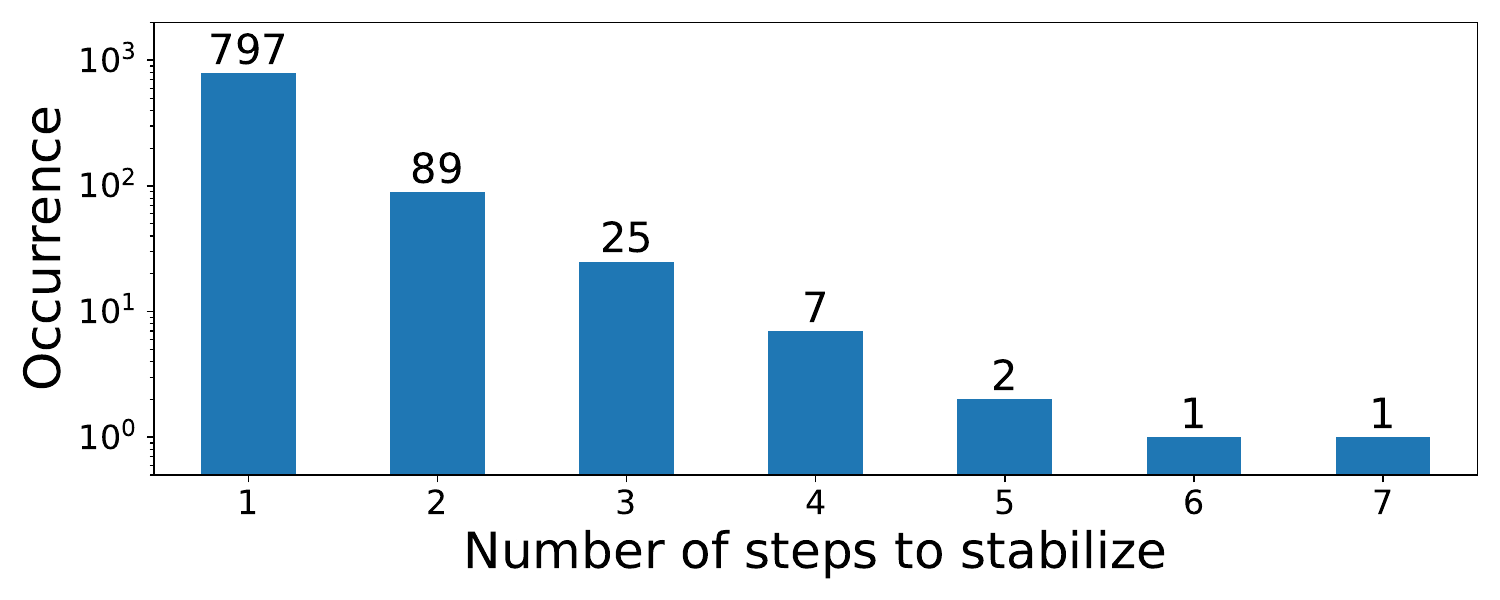}
  \caption{Number of occurrence of sustained gradient spikes vs contiguous length of appearance. Over 86\%, the spike lasts for only a single step.
  }
  \label{fig:gn-converge}
\end{figure}

Note that sustained spikes in the gradient norm leads to training divergence due to improper weight updates, even after gradient normalization through clipping (see Section\,\ref{ssec:train_hyper}). In Figure \ref{fig:gn-converge}, we show that the gradient spikes often last for a single step, and did not lead to training divergence. Specifically, we first track a running average ($r$) of gradient norm over a window of 20 steps to smooth out the natural fluctuations due to batching. We define occurrence of a gradient spike when the current gradient norm is higher than $r+0.1$. We then track the number of steps for gradient norm returning to less than $r+0.1$. Over 86\%, the spike deviates from running average for only a single step. 

Finally, we show the parameter $l_2$ norm in Figure \ref{fig:parameter_norm}. During first $\sim$250B tokens, the parameter norm increases consistently. This phase also coincides with the fast decreasing phase of training loss where model parameters converge from random initialization to a structured distribution. After that, the parameter norm consistently decrease since AdamW applies weight decay for regularization \cite{loshchilov2018decoupled}.

\subsection{Hardware and System Failures}
Pre-training on a cluster with thousands of accelerators often faces hardware-level errors that interrupts the training process. Those errors could be due to malfunction of AWS \trainium~chips or host machines, network communication timeouts, and so on \cite{borzunov2024distributed}. Manually restarting the pre-training from last-saved checkpoints demands significant developer time and causes low system uptime. For example, we performed an experimental training run (over 600 billion tokens) without automatic fault recovery, and we observed an average system uptime of 77.83\%. In \modelname~pre-training, we enabled automatic fault recovery mechanism in NxDT (see Section \ref{sec:back}), which automatically replaces the faulty nodes and restart training from latest checkpoints. The overall system uptime of \modelname~pre-training is then improved 20\% to 98.81\%. 

\subsection{Training Convergence and Instability}\label{sec:sr_err}
We describe a few changes we made before and during the training process for convergence and training stability. 

\textbf{Initialization}: We use a scaled initialization strategy for initializing model parameters. Specifically, the initial standard deviation of output layers in attention blocks and MLP layers are scaled by $1/\sqrt{2l}$ where $l$ is the layer index. Similar as discussed in \cite{takase2023spike}, we found better numerical stability and convergence with smaller initial variance on deeper layers.

\textbf{Normalization}: We used tensor parallelism to shard the model parameter matrices except normalization layers. For \modelname-7B, the normalization layer weights, however, are slightly different across TP ranks due to stochastic rounding. Empirically, we found the differences are small, and RMSNorm weights values are all close to 1. Note that with standard mixed precision strategy, we do not see such weight difference for \modelname-70B.


\change{\textbf{Checkpoint Averaging}: For \modelname-70B, the final checkpoint used for evaluation is an average of last two checkpoints in the training process. Similar as \cite{llama3, izmailov2019averagingweightsleadswider}, we found averaging the last two checkpoints provides better performance than each single checkpoint. We provide more detailed discussion in Section \ref{sec:ckpt_avg}.}

\textbf{Neuron Persistent Cache on Local Worker}: In \modelname~pre-training, all instances share a same file system using Amazon FSx\footnote{{https://aws.amazon.com/fsx/}} for storing data, checkpoints, logs, etc. However, we found that storing Neuron Persistent Cache\footnote{{https://awsdocs-neuron.readthedocs-hosted.com/en/latest/general/arch/neuron-features/neuron-caching.html}} on FSx may cause communication bottleneck because those cached graphs are frequently accessed by all \trainium~devices in the cluster. Such bottleneck may lead to communication timeout and affects training stability. Therefore, we instead store caches in file system of each local worker. 

\section{Evaluation} \label{sec:eval_tasks}
\textbf{Baselines:} We evaluate \modelname~against several open-source benchmark models. Since \modelname~structure is similar as LLaMA model, we include LLaMA-1 (7B, 13B, 65B) \cite{llama1}, LLaMA-2 (7B, 70B) \cite{llama2}, OpenLLaMA-1 (7B, 13B) and OpenLLaMA-2 (7B) \cite{openlm2023openllama}. The model architecture and composition of the training data of the models being compared are listed in Table \ref{tab:models}. OpenLLaMA-1 model is trained on RedPajama \cite{together2023redpajama} dataset. OpenLLaMA-2 model shares same structure as OpenLLaMA-1, but is trained on a different data mixture which includes data from Falcon-RefinedWeb \cite{refinedweb}, StarCoder \cite{starcoder}, and RedPajama \cite{together2023redpajama}. 

\begin{table}[ht]
\centering
\caption{Model architectures comparison.}
\label{tab:models}
\begin{tabular}{llcc}
\toprule
Model      & Sizes &  Sequence length  \\
\midrule
OpenLLaMA1 &7B, 13B   &  2048              \\
OpenLLaMA2 &7B   &  2048                \\
LLaMA1     &7B, 13B, 33B, 65B   &  2048              \\
LLaMA2     &7B, 13B, 70B   &  4096               \\
HLAT       &7B, 70B   &  4096             \\
\bottomrule
\end{tabular}
\end{table}


\begin{table*}[!h]
\centering
  \caption{\change{Evaluation of \modelname~against 4 open-source models on 6 groups of tasks described in Section \ref{sec:eval_tasks}. Numbers in the parentheses represent standard deviation, if available.}
}
  \label{tab:final_models}
\begin{tabular}{llllllllll}
\toprule
Model       & Size & MMLU       & CR         & WK         & RC         & Math       & \multicolumn{2}{c}{Code} & Average \\
-      &    -        & accuracy       & accuracy     &  exact match     &  accuracy     &  accuracy    & pass@1     & pass@10     &      -   \\
\midrule
OpenLLaMA-1 & 7B         & 30.5       & 58.4       & 40.6       & 70.5       & 5.2        & 4.5        & 13.4        & 41.2    \\
OpenLLaMA-2 & 7B         & 41.1       & 61.3       & 37.9       & 72.4       & 6.8        & 9.7        & 25          & 44.9    \\
LLaMA-1     & 7B         & 35.1       & 63.5       & 43.6       & 76.5       & 11         & 10.5       & 21.3        & 47.4    \\
LLaMA-2     & 7B         & 45.3       & 64         & 45.2       & 77.4       & 14.6       & 12.2       & 25.2          & 49.2    \\
\textbf{HLAT-7B}     & 7B         & 41.3 (3.6) & 59.5 (1.2) & 38.8 (0.5) & 72.5 (0.8) & 9.4 (0.8)  & 7.6        & 19.8        & 44.6    \\
\midrule
OpenLLaMA-1 & 13B        & 43.5       & 62       & 45.9       & 72.3       & 8.3        & 7        & 17        & 47.1    \\
LLaMA-1     & 13B        & 46.9       & 65.3       & 49.7       & 78.1       & 17.8       & 15.8       & 22.6        & 53.1    \\
LLaMA-2     & 13B        & 45.3       & 66.3       & 50.5       & 81.7       & 28.7       & 18.3       & 30.5        & 54.6    \\
LLaMA-1     & 33B        & 57.8       & 68.9       & 54.6       & 83.1       & 35.6       & 21.7       & 38.4        & 59.2    \\
\midrule
LLaMA-1     & 65B        & 63.4       & 69.8       & 57         & 85.3       & 50.9       & 23.7       & -           & 62.1    \\
LLaMA-2     & 70B        & 68.9       & 70.7       & 59         & 85         & 56.8       & 30.5       & 59.4           & 64.7    \\
\textbf{HLAT-70B}   & 70B        & 65.1 (3.4) & 67.3 (1.2) & 54.5 (0.6) & 82.6 (0.7) & 48.5 (1.4) & 21.4      & 57.9        & 60.8   \\
\bottomrule
\end{tabular}
\end{table*}

{\bf Evaluation Tasks:} 
We evaluate \modelname~against baselines on 7 groups of tasks including both zero-shot and few-shot tasks \cite{chang2023survey}. We use HumanEval \cite{humaneval} for coding tasks, and Language Model Evaluation Harness \cite{lmevalharness} for others. 

\textbf{Massive Multitask Language Understanding (MMLU)} \cite{hendrycks2021ethics,hendryckstest2021} contains 57 tasks, spanning STEM, social sciences, humanities, and other subjects. The difficulty ranges from elementary to professional levels. The breadth of the dataset tests model's overall problem solving and knowledge ability. 


\textbf{Commonsense Reasoning (CR)} consists of 6 datasets: PIQA \cite{bisk2020piqa}, HellaSwag \cite{zellers2019hellaswag}, WinoGrande \cite{sakaguchi2021winogrande}, ARC easy and challenge \cite{ARC}, and OpenBookQA \cite{openlm2023openllama}. These multi-choice tasks include carefully crafted riddles, puzzles, and scenarios designed to probe a model's ability to leverage implicit knowledge, make logical inferences, and navigate the rules of physical and social worlds. 

\textbf{World Knowledge (WK)} includes NaturalQuestions \cite{natualquestions} and TriviaQA \cite{joshi2017triviaqa}. Both tasks are designed to test model's question-answering ability in \textit{closed book} setting. The models are not provided documents that may contain information about the question, and it has to rely on information learnt or memorized in pre-training data. 

\textbf{Reading Comprehension (RC)} uses BoolQ \cite{clark2019boolq} to test model's \textit{open book} comprehension ability. BoolQ is a question answering dataset for yes/no questions. Each example is a triplet of \texttt{(question, passage, answer)}, with the title of the page as optional additional context. The model is required to answer the question based on the given context in \texttt{passage}. 

\textbf{Math} ability is evaluated with GSM8K (Grade School Math 8K) \cite{cobbe2021gsm8k}. GSM8K contains 8,500 grade school math problems. Both problems and answers are provided in natural language. These problems take between 2 and 8 steps to solve, which is ideal for testing basic multi-step reasoning ability.

\textbf{Code} evaluation uses HumanEval \cite{humaneval} dataset including 164 programming problems with a function signature, docstring, body, and several unit tests. They were handwritten to ensure not present in the training set of the models. 

\subsection{Performance against open-source Models}
We compare the performance of \modelname~with other open-source benchmarks in Table \ref{tab:final_models}. The numbers are reported in percentage and for \modelname~results, we include both mean and standard deviation (in the parentheses, if available). We also report an average score over all tasks in the last column. 

\change{\modelname-7B performs better than OpenLLaMA-1 and is on-par with OpenLLaMA-2. Both \modelname-7B and OpenLLaMA models have some gap with LLaMA-1 and LLaMA-2, which is likely due to the training data quality.
Even though the data composition of RedPajama-1T is similar as those used in LLaMA-1, the data cleaning pipeline and final data quality are different, which therefore affects the model performance \cite{gunasekar2023textbooksneed}. 
For \modelname-70B, we use the same training dataset as the 7B model for consistency. Although there is no OpenLLaMA baseline for a fair comparison, \modelname-70B performs better than LLaMA-1 and LLaMA-2 models of smaller sizes. The model performance gap with LLaMA-1 (65B) and LLaMA-2 (70B) is also smaller than those on 7B models. We acknowledge the lack of effort on data quality improvement, but our main goal is to showcase the effectiveness and efficiency of AWS \trainium.
}

On MMLU (5-shot), both \modelname~models perform better than OpenLLaMA-1 and LLaMA-1 models of similar size. The performance is slightly worse than LLaMA-2 family of models, likely due to the difference in training dataset size and composition \cite{llama1}. 

On Commonsense Reasoning (0-shot) and World Knowledge (5-shot), \modelname-7B performs similar to OpenLLaMA-1 and OpenLLaMA-2 models. By diving deep into performance on each individual task, \modelname-7B excels in 19/29 tasks as compared with OpenLLaMA-1, and 15/29 tasks compared with OpenLLaMA-2. Both \modelname~and OpenLLaMA models have some gaps with LLaMA-1 and LLaMA-2 models, which may be due to the training set quality. Nevertheless, the gap ($\sim3\%$) is consistent on 7B and 70B models. 

On Math problems (GSM8K, 8-shot), \modelname-7B performs significantly better than OpenLLaMA-1 and OpenLLaMA-2. As will be discussed in the next section, \modelname-70B has a big improvement of Math ability in later training stage. \modelname-70B performs similar as LLaMA1-65B, and we observed significant improvement in upsampling training stage. 

On Coding problems, both \modelname-7B and \modelname-70B perform comparable with LLaMA-1. \modelname-7B performs better than OpenLLaMA-1 and worse than OpenLLaMA-2 and LLaMA-2. First, for OpenLLaMA-1, the tokenizer merges consecutive spaces which negatively affects the coding performance, as it eliminates important information such as indentation and line breaks. This issue is subsequently fixed in OpenLLaMA-2, which explains its better performance. Besides, OpenLLaMA-2 is trained with additional code data from StarCoder which also contributes to performance improvement. 

\subsection{Intermediate Model Performance}
\begin{figure*}[!ht]
    \centering
    \begin{minipage}{0.55\columnwidth}
    \centering
        \includegraphics[width=0.9\linewidth]{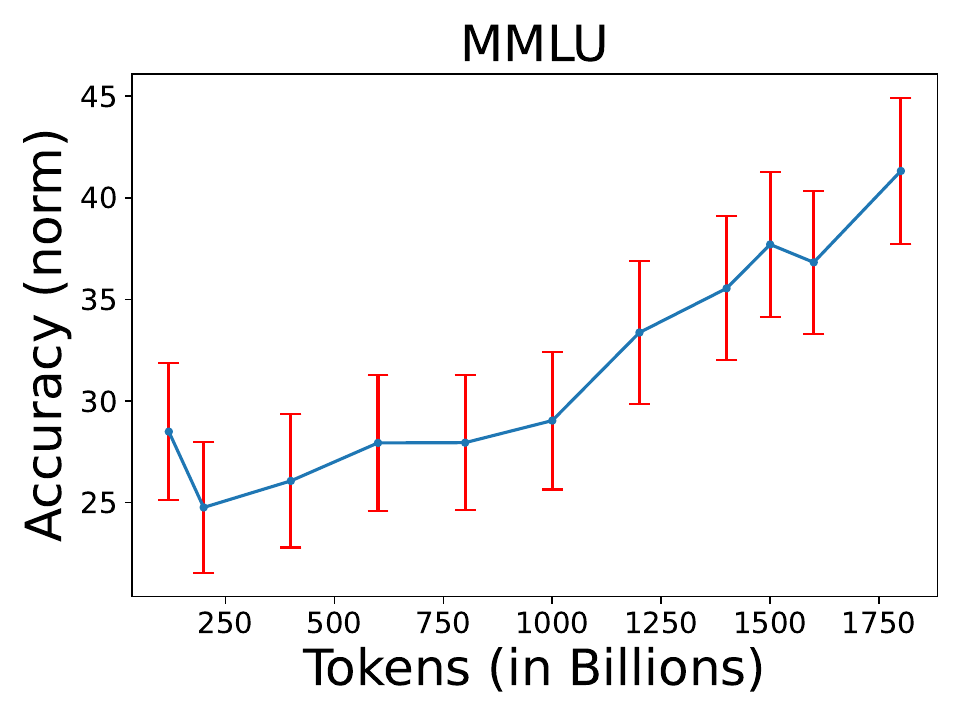}
        \subcaption{}
    \end{minipage}
    \begin{minipage}{0.55\columnwidth}
    \centering
        \includegraphics[width=0.9\linewidth]{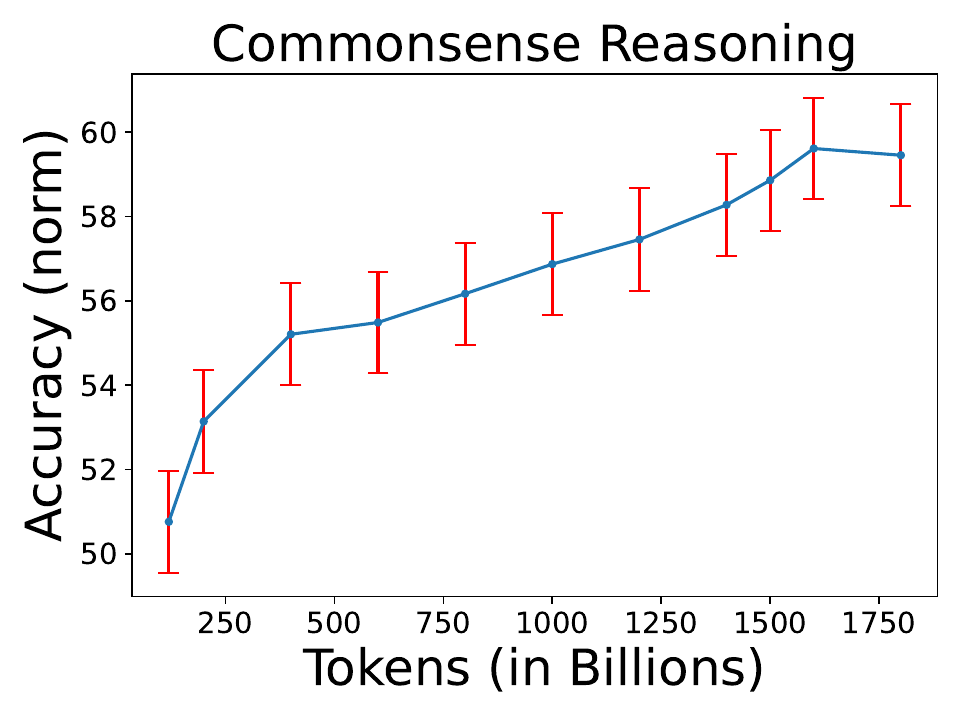}
        \subcaption{}
    \end{minipage}
    \begin{minipage}{0.55\columnwidth}
    \centering
        \includegraphics[width=0.9\linewidth]{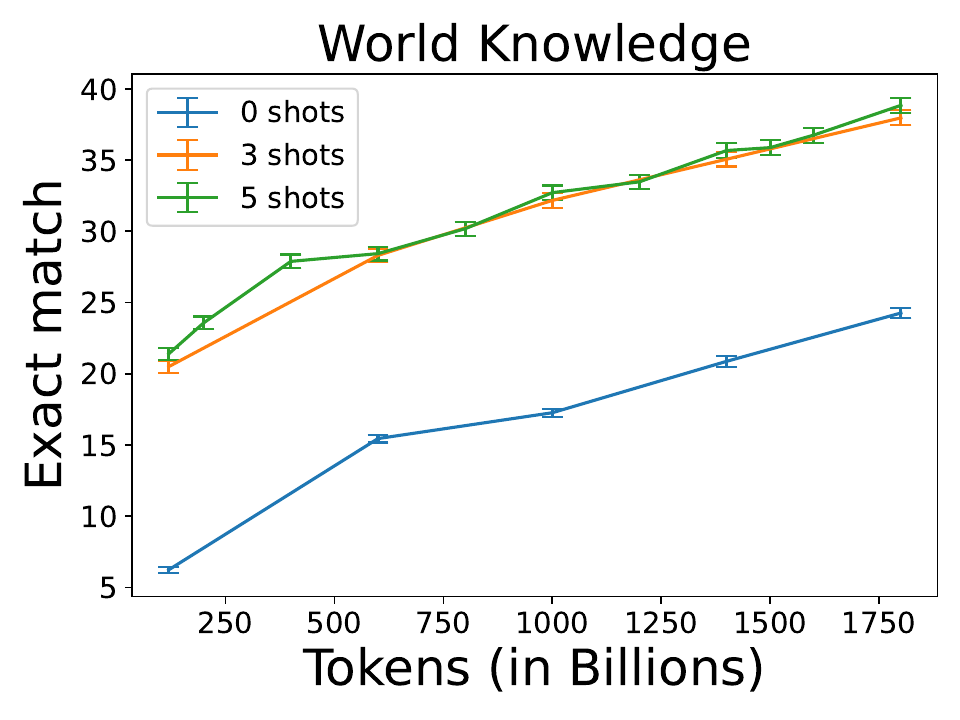}
        \subcaption{}
        \label{fig:wk}
    \end{minipage}
    \begin{minipage}{0.55\columnwidth}
    \centering
        \includegraphics[width=0.9\linewidth]{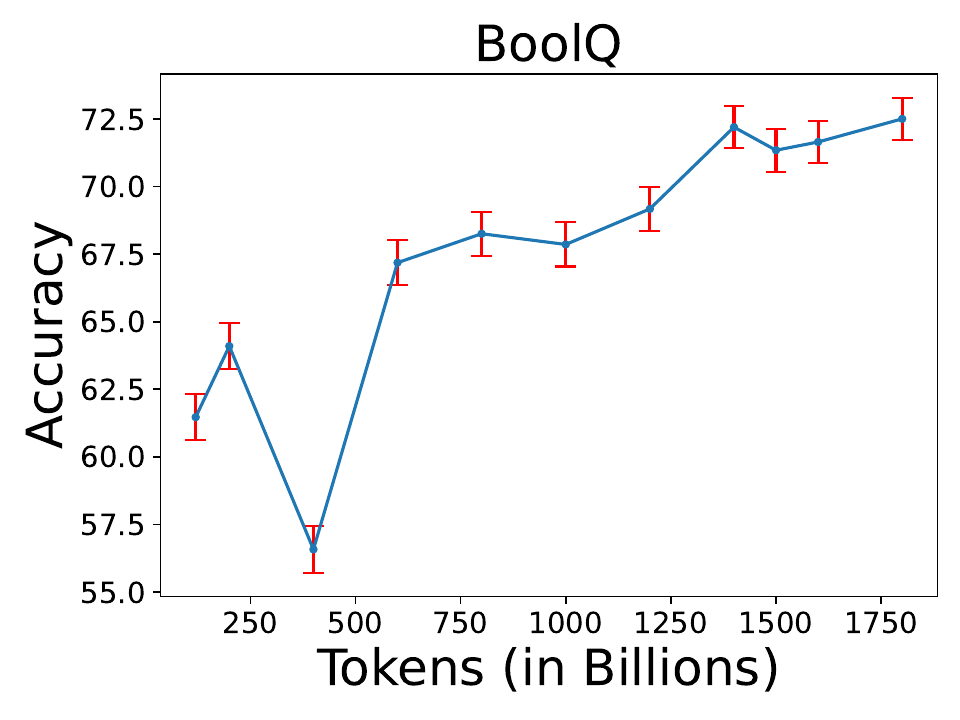}
        \subcaption{}
    \end{minipage}
    \begin{minipage}{0.55\columnwidth}
    \centering
        \includegraphics[width=0.9\linewidth]{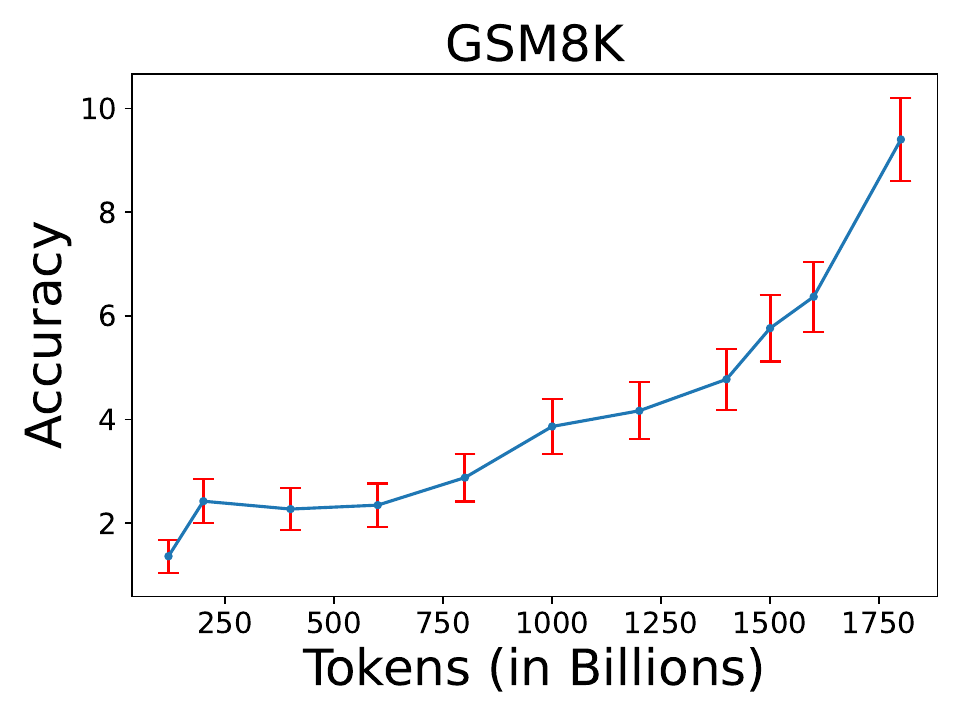}
        \subcaption{}
        \label{fig:math}
    \end{minipage}
    \begin{minipage}{0.55\columnwidth}
    \centering
        \includegraphics[width=0.9\linewidth]{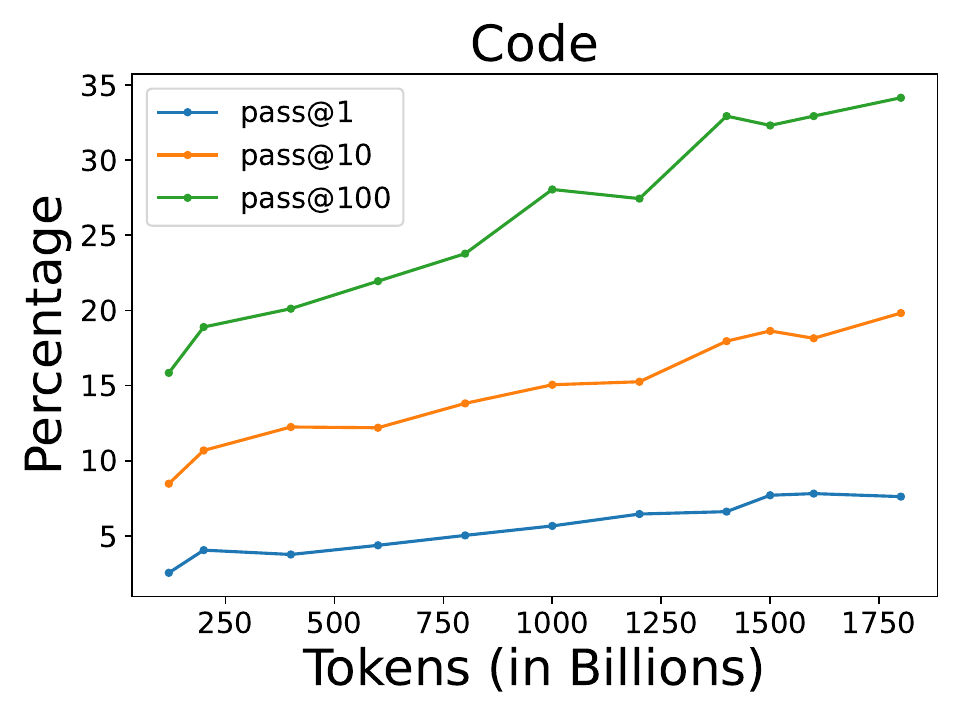}
        \subcaption{}
        \label{fig:code}
    \end{minipage}
    \caption{Intermediate model performance with number of seen tokens for \modelname-7B.}
    \label{fig:intermediate-7B}
\end{figure*}

\begin{figure*}[!t]
    \centering
    \begin{minipage}{0.55\columnwidth}
    \centering
        \includegraphics[width=0.9\linewidth]{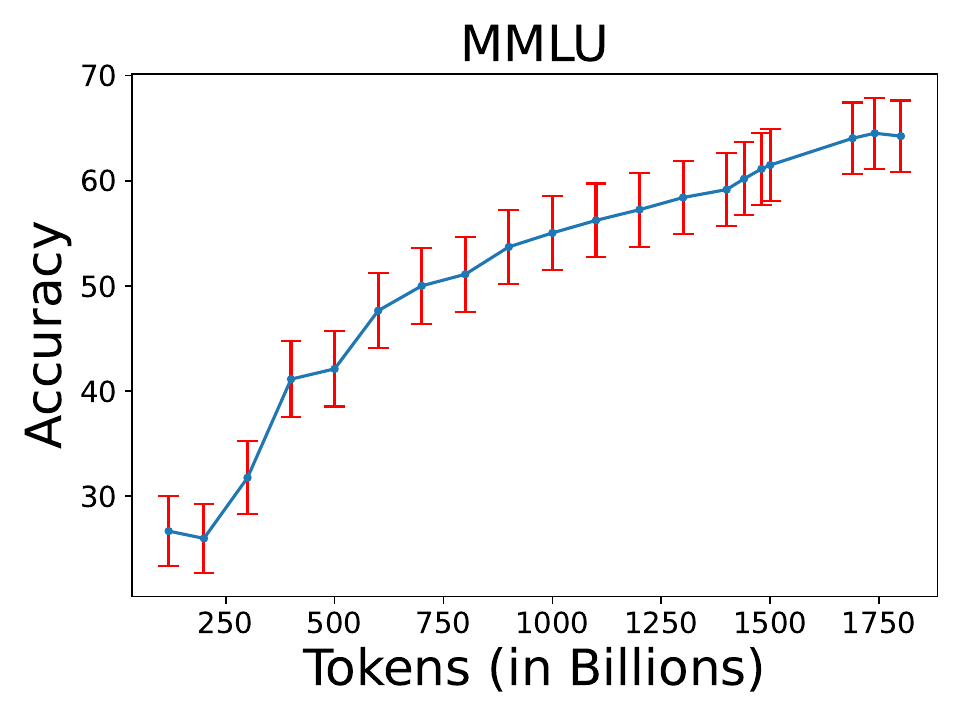}
        \subcaption{}
    \end{minipage}
    \begin{minipage}{0.55\columnwidth}
    \centering
        \includegraphics[width=0.9\linewidth]{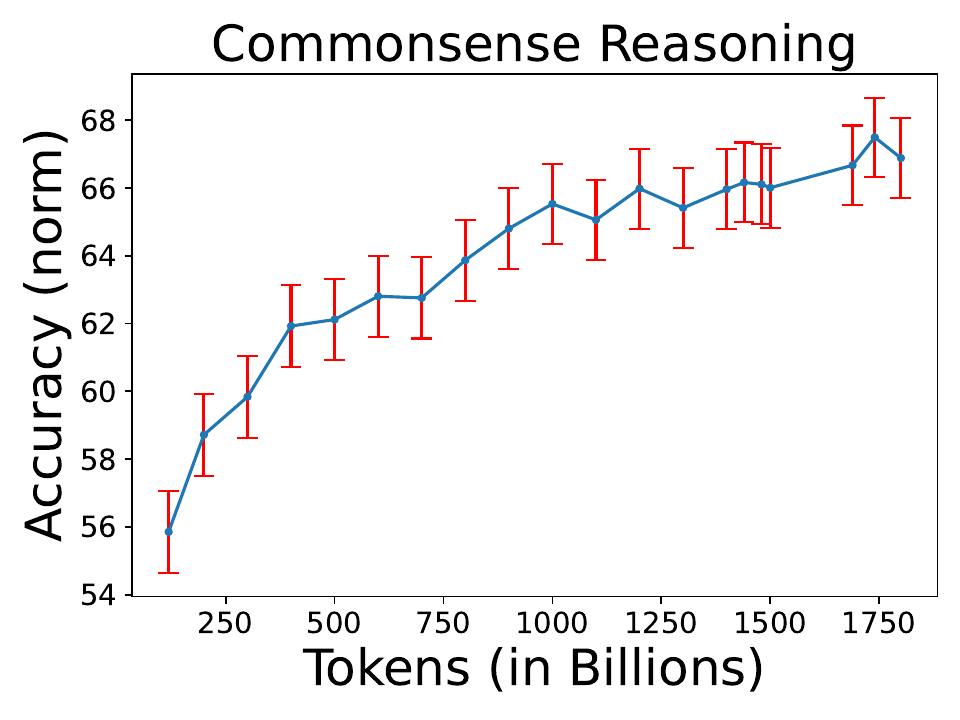}
        \subcaption{}
    \end{minipage}
    \begin{minipage}{0.55\columnwidth}
    \centering
        \includegraphics[width=0.9\linewidth]{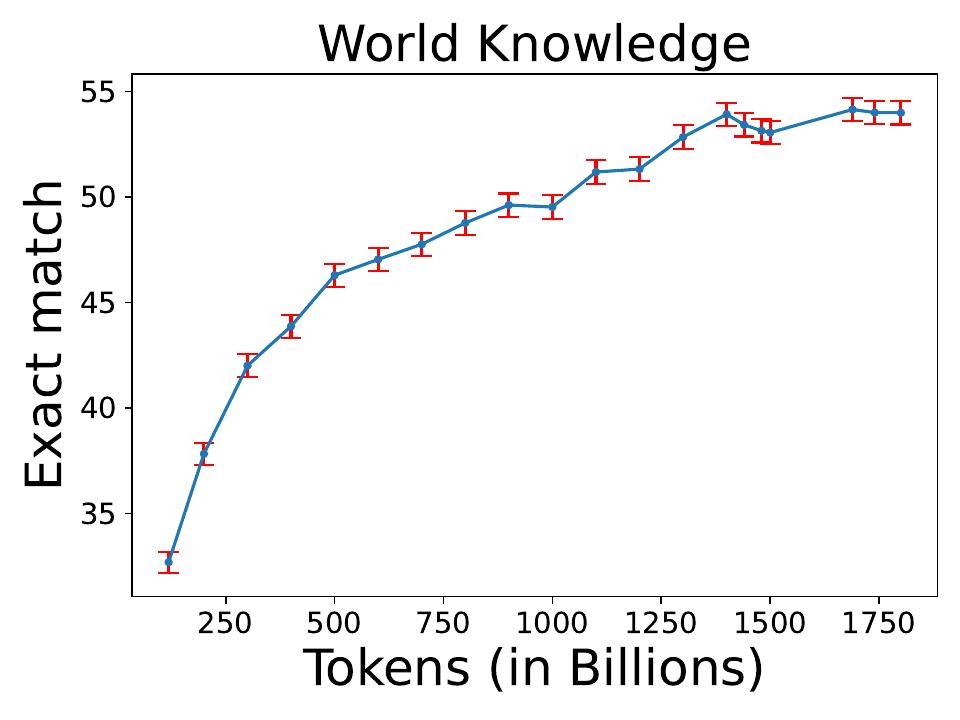}
        \subcaption{}
    \end{minipage}
    \begin{minipage}{0.55\columnwidth}
    \centering
        \includegraphics[width=0.9\linewidth]{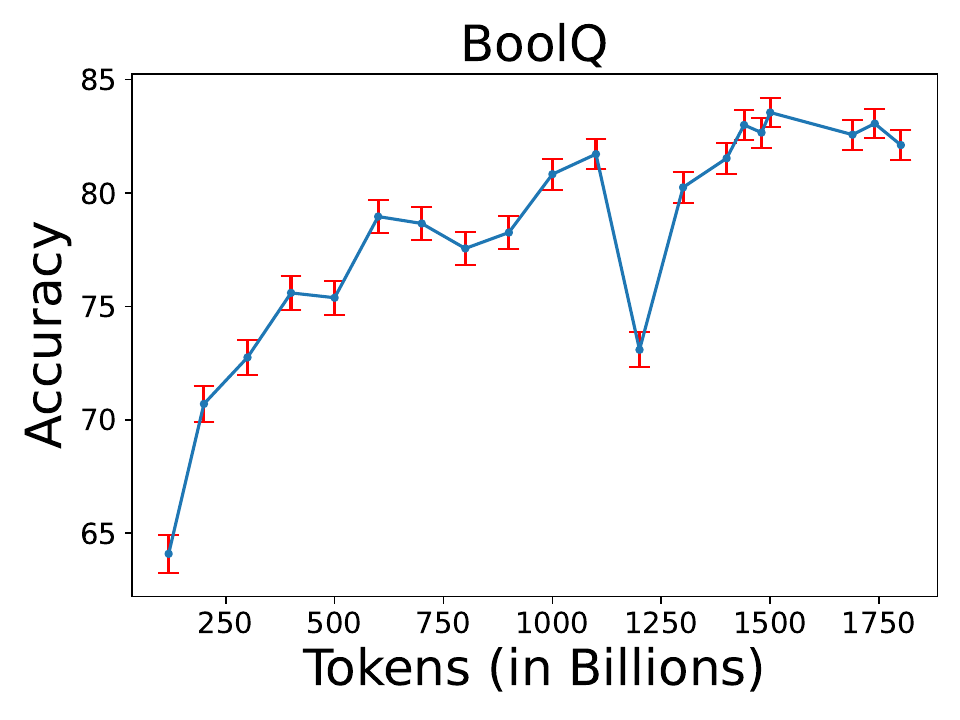}
        \subcaption{}
    \end{minipage}
    \begin{minipage}{0.55\columnwidth}
    \centering
        \includegraphics[width=0.9\linewidth]{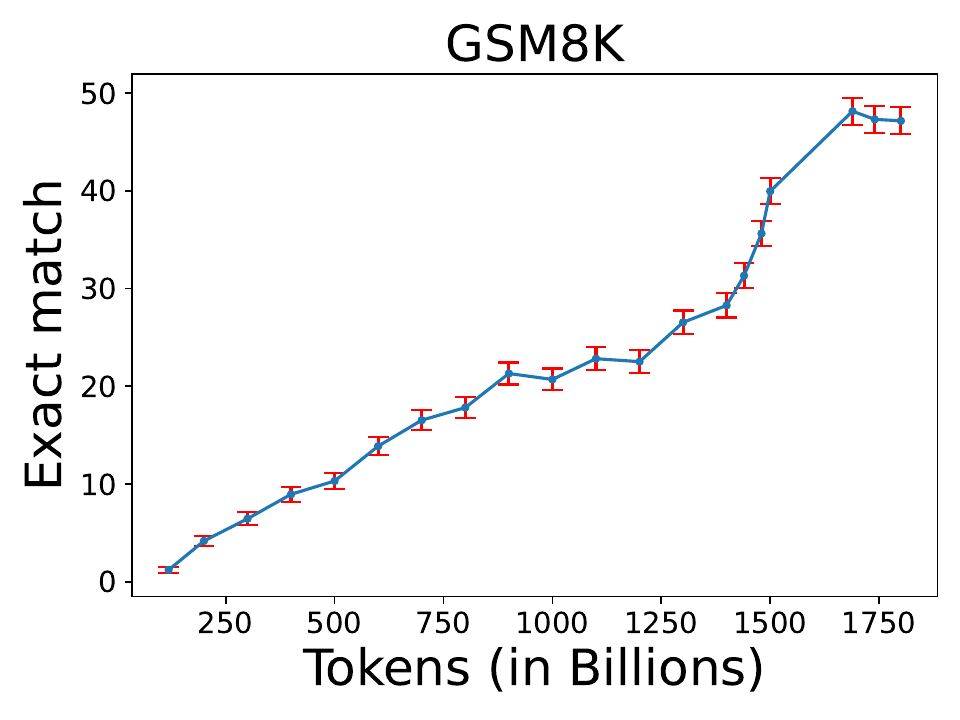}
        \subcaption{}
    \end{minipage}
    \begin{minipage}{0.55\columnwidth}
    \centering
        \includegraphics[width=0.9\linewidth]{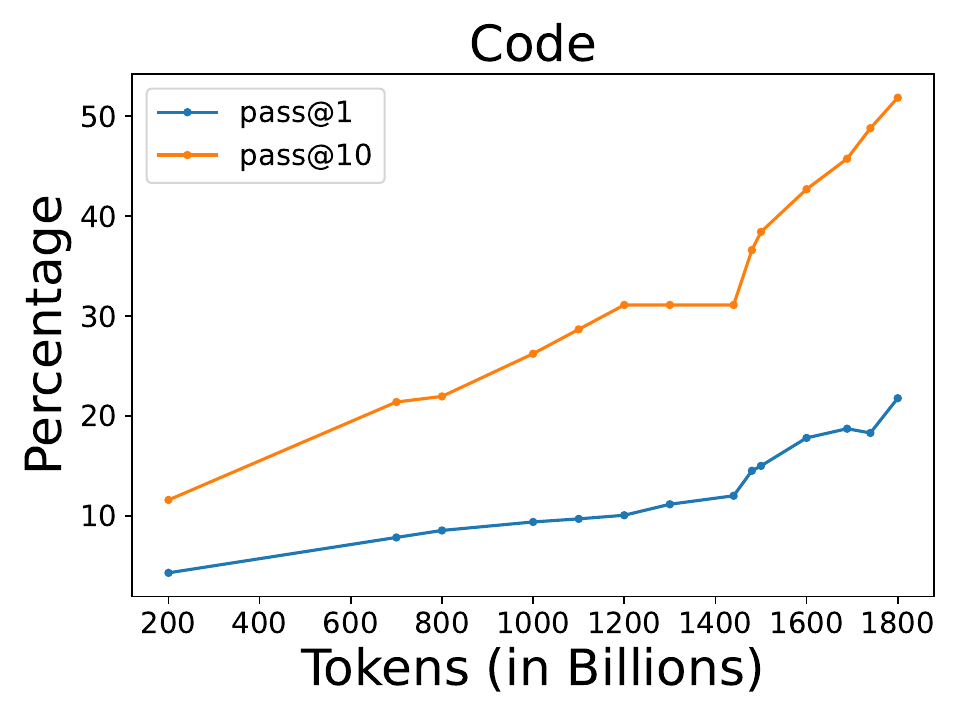}
        \subcaption{}
    \end{minipage}
    \caption{Intermediate model performance with number of seen tokens for \modelname-70B.}
\label{fig:intermediate-70B}
\end{figure*}

During the model training, we also evaluate the intermediate checkpoints about every 200 billion tokens. Figure \ref{fig:intermediate-7B} and Figure \ref{fig:intermediate-70B} show the model performance of \modelname-7B and \modelname-70B with respect to number of seen training tokens (in billions), respectively. On most benchmarks, the performance improves steadily, and correlates with the training loss. 

We found that for different tasks, the model converges at different rates. For Commonsense Reasoning, the model accuracy improves quickly at beginning of training, and starts to saturate at later training stages. This is similar as the trends observed in other LLM model trainings \cite{llama1,groeneveld2024olmo}. 

However, for Math task (GSM8K) shown in Figure \ref{fig:math}, the learning curve shows an exponentially increasing trend. It increase very gradually for the initial $\sim$1 trillion tokens and begins to improve significantly during the later stages of training. Intuitively, this seems to indicate that the model is able to grasp more logical abilities after entering a relatively stable training plateau. We defer further research into this behavior as a future work. 

For World Knowledge task shown in Figure \ref{fig:wk}, the performance increases almost linearly with number of training tokens. Since this is a \emph{closed book} test and mainly evaluates the model's ability of memorizing facts in pre-training data, the model seems to consistently improve its ability on this domain with more training steps and epochs. In addition, we tested if the trending is related to number of shots used in evaluation. It turns out that the trends are very similar for zero-shot, 3-shot, and 5-shot tests.

Those observations indicate the necessity of a set of evaluation tasks covering a wide range of domains for LLM pre-training. A single validation set or evaluation tasks from narrow domains may not fully reflect the actual over- or under-fitting of the model for general downstream performance.

\subsection{Upsampling} \label{sec:upsampling}
\begin{table}[!ht]
\centering
\caption{Upsampling dataset composition for \modelname-70B.}
\label{tab:us}
\begin{tabular}{llll}
\toprule
                          & Datasets      & \multicolumn{1}{l}{Size} & \multicolumn{1}{l}{Percentage} \\
& & (billions of tokens) & \\
\midrule
\multirow{2}{*}{Web Data} & Wikipedia\cite{together2023redpajama}     & \multirow{2}{*}{90}                  & \multirow{2}{*}{35.47\%}       \\
                          & C4\cite{together2023redpajama}            &                                      &                                \\
\midrule
\multirow{4}{*}{\begin{tabular}[c]{@{}l@{}}Domain \\ Specific\end{tabular}} & StackExchange & \multirow{4}{*}{104.7} & \multirow{4}{*}{41.27\%} \\
                          & Arxiv\cite{together2023redpajama}         &                                      &                                \\
                          & Open-Web-Math\cite{paster2023openwebmath} &                                      &                                \\
                          & PeS2o\cite{peS2o}         &                                      &                                \\
\midrule
Code                      & Github\cite{together2023redpajama}        & 59                                   & 23.26\%                        \\
\midrule
Total                     &    -           & 253.7                                & 15.16\%     \\
\bottomrule
\end{tabular}
\end{table}
\change{
During \modelname-70B training, we upsampled the training dataset in last 400B tokens. Specifically, we use 35.47\% web data, 41.27\% math data, and 23.26\% coding data with more details listed in Table \ref{tab:us}. In Figure \ref{fig:intermediate-70B}, we plot the evaluation performance of \modelname-70B with seen training tokens. In upsampling training stage, that is, after 1400B tokens, we observe significant model performance improvement over math, coding, and MMLU performance. It improved math by 10\% and coding by 5\%. This is consistent with the findings in LLaMA-3 \cite{llama3}, where the researchers found significant improvement of LLaMA-3 8B model on math problems. However, they mentioned such method did not help much for 405B models. Our experiment fills the model size gap, and shows that upsampling still helps for a 70B model. 

\subsection{Checkpoint Averaging} \label{sec:ckpt_avg}
For \modelname-70B, we average the last two checkpoints used in pre-training to generate a checkpoint for final evaluation \cite{llama3,izmailov2019averagingweightsleadswider}. Table \ref{tab:ckpt_avg} compares the individual checkpoints with 1740B training tokens and 1800B training tokens, as well as the averaged checkpoints. The averaged checkpoints outperforms individual checkpoints on average performance, as well as some individual tasks. }
\begin{table}[!ht]
\centering
  \caption{Evaluation of \modelname-70B with individual and averaged checkpoints.}
  \label{tab:ckpt_avg}
\begin{tabular}{llllllll}
\toprule
Checkpoint    & MMLU & RC   & WK   & CR   & Math & Code & Avg. \\
\midrule
1740B   & 64.5 & 67.5 & 54   & 83.1 & 47.3 & 18.3 & 60.3    \\
1800B   & 64.2 & 66.9 & 54   & 82.1 & 47.2 & 21.8 & 60.2    \\
Average & 65.1 & 67.3 & 54.5 & 82.6 & 48.5 & 21.4 & 60.8   \\
\bottomrule
\end{tabular}
\end{table}

\subsection{Truthfulness and Bias}
We report the model's truthfulness and bias using TruthfulQA \cite{lin2022truthfulqa} and CrowS-pairs \cite{nangia2020crows}. TruthfulQA presents a collection of meticulously crafted questions spanning diverse domains such as health, law, finance, and even politics. These queries deliberately target areas where human intuition and personal biases can lead to incorrect responses, and measure an LLM's resistance to misinformed or erroneous knowledge. CrowS-Pairs is a benchmark designed to probe LLMs for social biases across nine categories, including gender, religion, race/color, sexual orientation, age, nationality, disability, physical appearance and socioeconomic status. Each example is composed of a stereotype and an anti-stereotype. 
\begin{table}[!ht]
\centering
  \caption{Model Truthfulness and Bias evaluation. CrowS-pairs (CSP) uses \textit{percentage of stereotypes} as metric and TruthfulQA (TQA) uses \textit{multiple choice accuracy} as metric.}
  \label{tab:bias}
\begin{tabular}{llllll}
\toprule
Dataset     & Size & CSP ($\downarrow$) & CSP & TQA ($\uparrow$)       & TQA        \\
Tasks     & -    & english     & french      & mc1 & mc2 \\
\midrule
OpenLLaMA-1 & 7B   & 64.6        & 50.1        & 23.1              & 35.1              \\
OpenLLaMA-2 & 7B   & 65.6        & 51.7        & 22.6              & 34.6              \\
LLaMA-1     & 7B   & 53.7        & 47.5        &   22.0                &  34.1                 \\
LLaMA-2     & 7B   & 66.9        & 54.9        &     25.2              &      39.0             \\
\textbf{HLAT-7B}        & 7B   & 65.2        & 54.5        & 23.6              & 37.2              \\
\midrule
LLaMA-1     & 65B  & 69.3        & 58.3        & 27.9                & 42.6                  \\
LLaMA-2     & 70B  &  69.8           & 63.5            &  30.6                 &  44.8                 \\
\textbf{HLAT-70B}        & 70B  & 68.1        & 59.1        & 32.3              & 45.9      \\       
\bottomrule
\end{tabular}
\end{table}

We present the results in Table \ref{tab:bias} with 0 shot inference. For TruthfulQA, we measure the multiple-choice score, and higher score shows better truthfulness. For CrowS-Pairs, it measures the percentage of models choosing answers of stereotypes, so lower scores indicates smaller bias. Overall, \modelname~performs similar to other open-source models.

\subsection{Efficiency and Scalability}
We describe the training efficiency in terms of Cost per 4-million tokens (CPT) and scalability reported in \cite{isac2024trnpaper}. The CPT is defined as $\text{CPT} = \frac{C}{T\times 3600} \times 4\mathrm{e}^6$, where $C$ is instance cost per hour (\$21.50 for Trainium, and \$32.77 for GPU), $T$ is training throughput (tokens per second). CPT quantifies both the training speed and also hardware cost. We use this metric to compare training efficiency of Trainium and GPU. 
\begin{figure}[!thb]
  \centering
  \includegraphics[width=.85\linewidth]{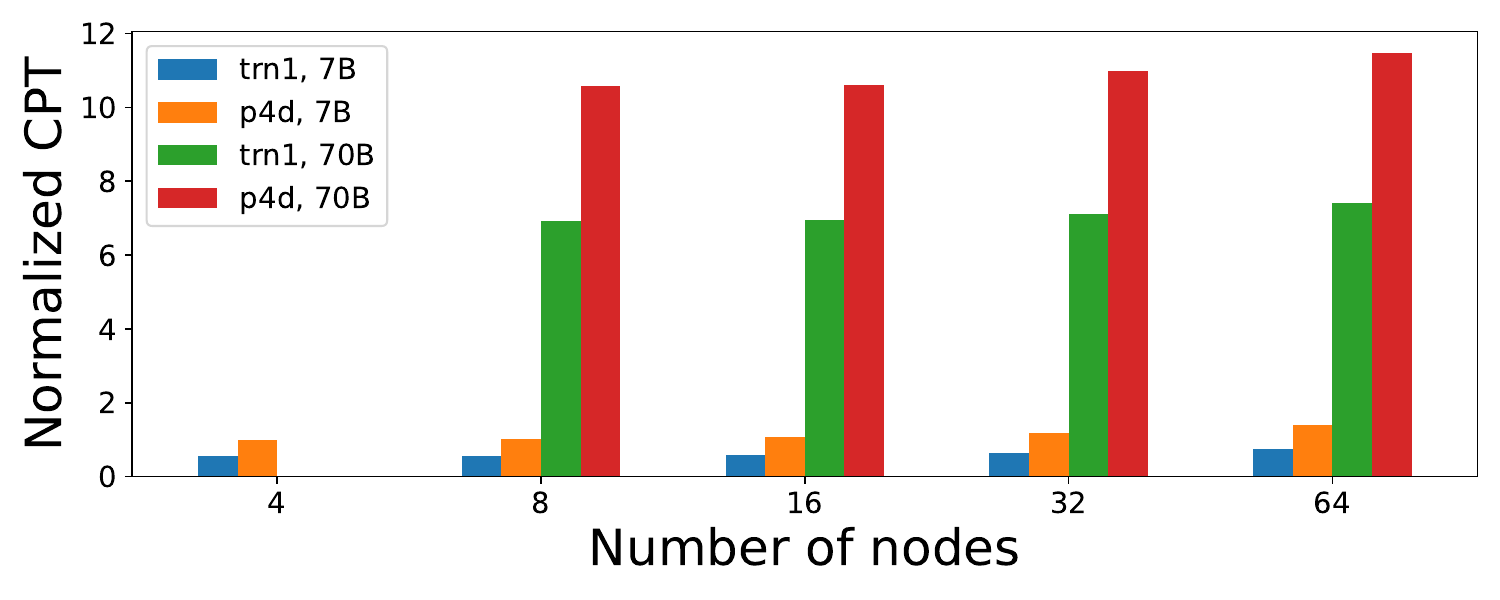}
  \caption{Normalized cost per 4 million tokens (CPT) for 7B and 70B models on AWS \trainium~with various number of nodes. CPT of GPU baseline (\emph{p4d}, 7B) with 4 nodes is normalized to 100\%. 70B models ran into out-of-memory on 4 nodes.}
  \label{fig:cost}
\end{figure}

\change{For comparison, the GPU baseline is established using \emph{p4d.24xlarge} instances and NeMo 23.08 \cite{Harper_NeMo_a_toolkit} (available inside NeMo docker container with tag \texttt{23.08}) software stack. Figure \ref{fig:cost} plots the normalized CPT of training on \trainium~and scaling. The \trainium~CPT is normalized, such that the CPT of the GPU baseline (\emph{p4d}, 7B) on 4 nodes is 100\%. Overall, the training cost on \emph{trn1} is approximately 60\% of GPU, and is consistent with the number of nodes. In addition, the CPTs on 70B models are roughly 10 times of those on 7B models. }

\subsection{Model Limitation}
We note some limitations of \modelname~in this section. Similar as other LLMs, \modelname~suffers a set of limitations such as hallucinations, potential non-factual generations, biases, and toxicity \cite{zhang2023hallucination}. For example, although comparable with other open-source pre-trained models, the bias of \modelname~is still relative high on some subjects such as sexual orientation, physical appearance, religion, and socioeconomic (see Table \ref{tab:bias}). This is partially due to the usage of publicly available datasets. More importantly, as a pre-trained model, \modelname~has not gone through a supervised finetuning and human preference alignment. Those fine-tuning methods have been shown to be able to alleviate some limitations of pre-trained LLMs \cite{llama2}. Another limitation is that our training is stopped after 1.8 trillion tokens. As is suggested by LLaMA-3 \cite{llama3}, \modelname~may be able to further improve on certain tasks, such as math, world knowledge, MMLU, and coding, with more training tokens.

\section{Best Practices \& Future Directions}
In this section, we share some best practices we observed for training on AWS \trainium, and raise open questions for future research. 

\textbf{Parallelism}:
NxDT supports TP up to 32 degrees and pipeline parallelism. For a 7B model, we found that the combination of TP=8 and PP=1 provides the highest training throughput, but not for \modelname-70B. So  the optimal parallelism configuration varies with model sizes and architectures. To achieve the highest training throughput, parallelism configuration needs to be jointly optimized with choice of activation checkpointing method, gradient accumulation steps, and training precision, to balance memory and communication costs.

{\bf Training Precision}: NxDT supports various training precision configurations, including full precision (FP32), BF16 with and without SR, standard mixed precision training, etc. Full precision training is often memory-wise infeasible for multi-billion LLMs. We compared multiple training strategies for \modelname: pure BF16, BF16 with SR and standard mixed precision training. Empirically, we found that training loss of pure BF16 diverges. BF16 with SR shows similar training loss as mixed precision on \modelname-7B model, but converges slower on \modelname-70B. We finally chose BF16 with SR for higher throughput on \modelname-7B, but standard mixed precision on \modelname-70B. For models of other sizes and architecture, warm-up study may be needed to decide the optimal training precision. Usually, the divergence can be observed in first few thousands of steps. 

\textbf{Choice of $\beta_2$}: We observed that using $\beta_2=0.99$ causes training instability and slower convergence. This is related to the choice of BF16 with SR training precision. A large $\beta_2$ fails to capture the gradient explosion at current and recent steps, and hence does not effectively reduce the gradients in occurrence of gradient explosion. Switching to $\beta_2=0.95$ addresses the above-mentioned problem. 

\textbf{Weight decay}: We applied weight decay to all layers. Empirically, weight decay is not applied to normalization and bias layers \cite{devlin-etal-2019-bert}. In our experiment, we did not found much performance-wise difference of those two methods. 


\textbf{Pre-compilation}: \trainium~requires pre-compiling the scripts to graphs. The compilation takes some time, especially for large models. Debugging on training scripts (e.g., printing out intermediate tensors) may require re-compilation. Instead of directly developing on a large model, we found it more efficient to develop and test on a smaller model and scale up afterwards. 

\section{Related Work}
\textbf{LLM pre-training:} After the Transformer architecture \cite{vaswani2017attention} was introduced, BERT \cite{devlin-etal-2019-bert} was proposed to pre-train a language model on a large corpus of unlabeled data. Following the success of BERT model on various NLP tasks, many pre-trained language models are later introduced with different architectures and training methods, such as GPT-2 \cite{radford2019language}, RoBERTa \cite{liu2019roberta}, BART \cite{lewis-etal-2020-bart}, and so on \cite{LLMSurvey}. Studies later observed significant performance improvement of language models by increasing model size and training data \cite{hoffmann2022training}. Such abilities are further demonstrated in LLMs such as GPT-3 \cite{gpt3}, PaLM \cite{anil2023palm}, LLaMA \cite{llama1,llama2,llama3}, Falcon \cite{almazrouei2023falcon}, Gemini \cite{team2023gemini}, Phi \cite{gunasekar2023textbooksneed}, etc. Pre-trained on trillions of tokens, LLMs with tens or hundreds of billions parameters show remarkable ability in generating creative text contents, as well as a variety of downstream tasks, such as question answering, summarization, machine translation, programming, etc. \cite{LLMSurvey}.

\textbf{AI accelerators:} Most models are trained on NVIDIA GPU accelerators, such as GPT \cite{gpt3,OpenAI_GPT4_2023} and LLaMA \cite{llama1,llama2}. Falcon-180B \cite{almazrouei2023falcon} was trained on AWS SageMaker, with up to 4,096 A100 40GB GPUs using \emph{p4d} instances. However, the landscape of hardware accelerators for deep learning training has blossomed in recent years, with established players like NVIDIA GPUs facing fierce competition from custom offerings like Google's TPU and AWS \trainium. PaLM-2 \cite{anil2023palm} and OpenLLaMA \cite{openlm2023openllama} have demonstreated successful LLM pre-training on Google TPU. Recently, OLMo \cite{groeneveld2024olmo} is an open-source model developed by AI2. It has two models trained on AMD and Nvidia GPUs, separately. The two models have nearly identical performance on their evaluation suite by 2T tokens. AWS \trainium~is a machine learning accelerator developed for deep learning training with high performance and cost-competitiveness. Our work is the first demonstration of end-to-end multi-billion LLM pre-trained on AWS \trainium. Ultimately, the optimal choice depends on the specific needs of the training task, with further research required to fully explore the potential of each accelerator and their possible convergence in future architectures.

\section{Conclusion}
In this paper, we pre-train \modelname, a family of 7 billion and 70 billion parameter large language models, using AWS \trainium~over $\sim$1.8 trillion tokens. \modelname~follows the decoder-only architecture and is trained with up to 256 Amazon EC2 \emph{trn1.32xlarge} instances. We evaluate the performance of \modelname~against popular open-source baseline models including LLaMA and OpenLLaMA on a variety of popular benchmarking tasks. We find that \modelname~achieves model quality on par with these baseline models of similar sizes. This work demonstrates, for the first time, that AWS \trainium~with NxDT is able to successfully pre-train high-quality LLMs with high efficiency and low cost.

\bibliographystyle{IEEEtran}
\bibliography{IEEEabrv,refs}

\end{document}